\documentclass[letterpaper]{article} 
\usepackage{aaai23}  
\usepackage{times}  
\usepackage{helvet}  
\usepackage{courier}  
\usepackage[hyphens]{url}  
\usepackage{graphicx} 
\usepackage{natbib}  
\usepackage{caption} 
\usepackage{algorithm}
\usepackage{algorithmic}

\usepackage{paralist}
\usepackage[inline]{enumitem}
\usepackage{amsmath}
\usepackage{graphicx}
\usepackage{tabularx}
\usepackage{tikz}
\usepackage{xfrac}
\usepackage{xspace}
\usepackage{xcolor}
\usepackage{wrapfig}
\usepackage{bm}
\usepackage{makecell}
\usepackage[labelformat=simple]{subcaption}
\usepackage{float}
\usepackage{todonotes}
\usepackage{amsfonts}
\usepackage{booktabs}
\usepackage{balance}
\usepackage{longtable}
\usepackage{empheq}

\usepackage{newfloat}
\usepackage{listings}
\DeclareCaptionStyle{ruled}{labelfont=normalfont,labelsep=colon,strut=off} 
\floatstyle{ruled}
\newfloat{listing}{tb}{lst}{}
\floatname{listing}{Listing}
\def\eqref#1{equation~\ref{#1}}

\def\1{\bm{1}}

\DeclareMathAlphabet{\mathsfit}{\encodingdefault}{\sfdefault}{m}{sl}
\SetMathAlphabet{\mathsfit}{bold}{\encodingdefault}{\sfdefault}{bx}{n}

\newcommand{\valstd}[2]{$#1$ & $#2$}
\newcommand{\valstdb}[2]{$\mathbf{#1}$ & $#2$}

\title{Semiconductor Fab Scheduling with Self-Supervised and Reinforcement Learning%
\thanks{%
This work was 
funded by
KWF project 28472,
cms electronics GmbH,
FunderMax GmbH,
Hirsch Armbänder GmbH,
incubed IT GmbH,
Infineon Technologies Austria AG,
Isovolta AG,
Kostwein Holding GmbH, and
Privatstiftung Kärntner Sparkasse.}}
\author {
    Pierre Tassel,\textsuperscript{\rm 1}
    Benjamin Kovács,\textsuperscript{\rm 1}
    Martin Gebser,\textsuperscript{\rm 1}\textsuperscript{\rm 2}
    Konstantin Schekotihin,\textsuperscript{\rm 1}
    Patrick St\"ockermann,\textsuperscript{\rm 3}
    Georg Seidel\textsuperscript{\rm 3}
}
\affiliations {
    \textsuperscript{\rm 1} University of Klagenfurt\\
    \textsuperscript{\rm 2} Graz University of Technology\\
    \textsuperscript{\rm 3} Infineon Technologies\\
    pierre.tassel@aau.at, benjamin.kovacs@aau.at, martin.gebser@aau.at, konstantin.schekotihin@aau.at, patrick.stoeckermann@infineon.com, georg.seidel@infineon.com
}

\usepackage{bibentry}

\begin{document}

\maketitle

\begin{abstract}
    Semiconductor manufacturing is a notoriously complex and costly multi-step process involving a long sequence of operations on expensive and quantity-limited equipment.
    Recent chip shortages and their impacts have highlighted the importance of semiconductors in the global supply chains and how reliant on those our daily lives are.
    Due to the investment cost, environmental impact, and time scale needed to build new factories, it is difficult to ramp up production when demand spikes.
    This work introduces a method to successfully learn to schedule a semiconductor manufacturing facility more efficiently using deep reinforcement and self-supervised learning.
    We propose the first adaptive scheduling approach to handle complex, continuous, stochastic, dynamic, modern semiconductor manufacturing models.
    Our method outperforms the traditional hierarchical dispatching strategies typically used in semiconductor manufacturing plants, substantially reducing each order's tardiness and time until completion.
    As a result, our method yields a better allocation of resources in the semiconductor manufacturing process.
\end{abstract}

\section{Introduction} \label{sec:intro}
  
  Modern semiconductor manufacturing is one of the most complex industrial processes to date \citep{iwaiChallengesFutureSemiconductor2005, raviChipmakersAreRamping2021}.
  Typical production sequences involve a multitude of complex elementary operations handled by highly specialized 
  tools that require specific setups according to variable product and process designs
  \citep{yugmaIntegrationSchedulingAdvanced2015}.
  Over the years, Moore's Law has pushed the economic and manufacturing limits of 
  industry by approximately doubling the number of transistors per given area every two years.
  On the other hand, the costs for research and development, manufacturing, and testing become more and more elevated with each new chip generation \citep{ruppEconomicLimitMoore2011}.
  Thus, increasing semiconductor production capacities takes time, and it is impossible to ramp up chip output overnight \citep{raviChipmakersAreRamping2021, lesliePandemicScramblesSemiconductor2022}.
  Recent disruption in supply chains due to the Coronavirus disease (COVID-19) pandemic and severe droughts distorted the market for microchips, leading to a record of 22 weeks waiting time per order in October 2021, 
  rather than the around 12 weeks of late 2019, 
  where experts predict that supply will not catch up with demand for another year or more \citep{lesliePandemicScramblesSemiconductor2022}.
  In this regard, optimization of production processes is an important prerequisite for the mitigation of the current critical situation on the markets.
  Beyond the economic and manufacturing challenges, increased production facilities have a negative ecological impact, as
  the industry is very resource-intensive, especially in terms of electricity, land, and water consumption \citep{kuoAssessingEnvironmentalImpacts2022}.
  In 2010, the semiconductor industry accounted for $1.3$–$2.0\%$ of the total US electricity consumption in the manufacturing sector \citep{gopalakrishnanEnergyAnalysisSemiconductor2010}.
  The chip-making process is also water-intensive, and an average semiconductor factory consumes around $20\,000$ tons of water a day with an estimate of $97.26\%$ used during manufacturing \citep{frostSpatiallyExplicitAssessment2017}.
  Increasing semiconductor fab efficiency has thus been identified as one of the ways to achieve ecological benefits while reducing the environmental impact of semiconductor manufacturing \citep{agIndustrialApplicationsMaking}.
  
  Applications of machine learning techniques to combinatorial optimization problems like planning and scheduling have recently emerged as a prospective research area \citep{bengioMachineLearningCombinatorial2020}.
  As most of these problems are $\mathit{NP}$-complete or even harder, exhaustive search methods do not scale and quickly become inefficient.
  The same observation can be made for various scheduling problems, most of which are $\mathit{NP}$-hard \cite{DBLP:journals/mor/GareyJS76,SOTSKOV1995237}.
  Applying machine learning techniques to scheduling problems has shown to be a promising approach to generate good approximate solutions in reasonable time \citep{parkLearningScheduleJobshop2021,zhangLearningDispatchJob2020b, strickerReinforcementLearningAdaptive2018}.
  Recently, \citet{tasselReinforcementLearningDispatching2022} suggested an RL-based method that can handle very large instances of the resource constraint scheduling problem.
  However, previous work focuses on scenarios that significantly simplify real-world problems \cite{strickerReinforcementLearningAdaptive2018, tasselReinforcementLearningEnvironment2021a}. That is, considered problem instances represent finite, deterministic, and static scheduling problems with predefined sets of machines, jobs, and their operations, which remain constant during the scheduling process.
  Therefore, our work is the first to consider the \emph{continuous}, \emph{stochastic}, and \emph{dynamic} scheduling process of a semiconductor fab, addressing a wide 
  range of challenges encountered in large-scale production environments.

  \paragraph*{Contributions.} In this paper, we propose a novel adaptive scheduling method to improve the yield and reduce customer order delays in a realistic semiconductor manufacturing environment.
  Using \emph{Reinforcement Learning} (RL) and \emph{Self-Supervised Learning} (SSL), we train a single RL agent's \emph{Neural Network} acting as a global dispatcher. Our contributions can be summarized as follows:
  \begin{description}[leftmargin=5pt,labelindent=-5pt]
   \item \textbf{RL Training Approach.} Due to the specificities of the semiconductor manufacturing process and our model's architecture, we have developed a training loop using \emph{Self-Supervised Learning} to pre-train part of our network and derivative-free \emph{Evolution Strategies} to train another part of the network. 
   \item \textbf{Feature Set.} For each wafer lot, we have designed a feature set that can efficiently represent the current status of the lot in the manufacturing process.
   \item \textbf{Objective Function.} We have designed an objective function to compare the quality of different schedules. 
   This function aims to reduce the average tardiness and cycle-time, thus increasing the throughput of each lot type while incorporating their priorities. The function also considers unfinished lots (i.e., the WIP) in the computation of the score, correcting potential bias in evaluating a continuous production environment.
   \item \textbf{Model Architecture.} We introduce an order- and size-invariant neural network architecture for a single-agent dispatcher using global self-attention on the features of each wafer lot to be scheduled. 
   Our architecture can efficiently encode information such as the type of a required tool family without increasing the dimensions of the input representation.
   \item \textbf{Empirical Evaluation.} We evaluate our method on a large-scale, continuous, stochastic and dynamic semiconductor simulation model from the literature \citep{koppSMT2020SemiconductorManufacturing2020}, using an open source event-based semiconductor fabrication plant simulation \citep{kovacsCustomizableSimulatorArtificial2022}. 
   This testbed was designed to help researchers to assess the performance of integrated dispatching strategies and aims to credibly reflect the complexity of modern semiconductor manufacturing.
   It includes two models, one representing a High-Volume/Low-Mix (\textit{HV/LM}) production scenario and the other a Low-Volume/High-Mix (\textit{LV/HM}) production plan. 
   Our approach outperforms traditional dispatching heuristics in both of these scenarios, achieving so far unmatched state-of-the-art performance.
  \end{description}

\section{Background}  \label{sec:prelim}
The problem of scheduling in semiconductor manufacturing can be modeled as a variant of the Job-shop Scheduling Problem (JSP) \citep{guptaJobShopScheduling2006}.
In classical JSP, given $n$ jobs and $m$ machines, each operation of a job must be non-preemptively processed by exactly one machine within a minimal makespan, i.e., the total completion time for all jobs.
Although simple in appearance, the problem is $\mathit{NP}$-complete, and unless $\mathit{NP}=P$, it cannot be solved efficiently \citep{applegateComputationalStudyJobShop1991}.
Due to the limited practical applicability of the classical JSP, multiple variants of this problem, which better reflect real-world scheduling scenarios, have emerged over the years \citep{pottsFiftyYearsScheduling2009}.
For instance, in flow-shop scheduling, each job has exactly $m$ operations and an operation $i$ is non-preemptively executed on machine $i$.
In a flexible JSP, for every operation, there can be several machines able to process it. 
Complex JSP further extends the flexible variant with \textit{batching} machines, reentrant flows, sequence-dependent setup times and release dates \citep{knoppComplexJobShopScheduling}.
The term \textit{batching} is used in the context of parallel processing and/or in that of sequential processing (often referred to as \textit{cascading} or \textit{serial batching}),
denoting capacities for overlapping the execution times of multiple jobs' operations \citep{huangJobSchedulingCascading2017}.

However, a realistic model of the semiconductor manufacturing process cannot be directly mapped to existing JSP variants.
Therefore, \cite{koppSMT2020SemiconductorManufacturing2020} presented \textit{SMT2020}, a set of stochastic semiconductor fab simulation models, aiming to credibly represent the complexity of modern semiconductor manufacturing.
These models go beyond previously proposed semiconductor simulation testbeds such as MiniFab \citep{spierSimulationEmergentBehavior1995}, MIMAC \citep{hassounNewSimulationTestbed2017}, Harris \citep{kaytonFocusingMaintenanceImprovement1997}, and SEMATECH 300mm \citep{kibaSimulationFull300mm2009} by including larger-scale datasets and more realistic characterizations of manufacturing processes, incorporating the following features:
%

\begin{enumerate}[left=0pt]
  \item \textbf{Different Simulation Scenarios:} The simulation includes models of Low-Volume/High-Mix (\textit{LV/HM}) and High-Volume/Low-Mix (\textit{HV/LM}) wafer fabs, where mix stands for high/low variety of products, and volume for small/large quantities, respectively. 
  The \textit{LV/HM} scenario includes $10$ different products with $10$ associated routes,
  reflecting make-to-order (MTO) environments for supplying foundries, producing logic or memory devices.
  In the \textit{HV/LM} setting, $2$ high-volume products are considered to represent dedicated fabs, typically producing logic devices in a make-to-stock (MTS) environment.
  \item \textbf{Realistic Scale:} Each product has a dedicated route with at least $500$ steps, reflecting the complexity of modern wafer fabs.
  Similarly, both scenarios contain $106$ tool groups organized in $12$ families, sometimes called machine families or work centers.
  The \textit{LV/HM} scenario includes a total of $1\,265$ machines, while \textit{HV/LM} features $1\,071$ machines.
  At the beginning of the scheduling horizon, there are $2\,700$ lots in progress for the \textit{LV/HM} dataset and $3\,635$ for the \textit{HV/LM} scenario.
  \item \textbf{Machine Availability:} Simulation models include scheduled downtime due to planned tool maintenance and unscheduled downtime due to machine breakdown.
  \item \textbf{Batching and Cascading:} \textit{SMT2020} scenarios involve batch and/or cascade processing for certain tool groups. 
  \item \textbf{Loading and Transportation:} Loading and unloading times on the tools and transportation times between the different tool groups are considered in the simulation. 
  \item \textbf{Machine Setup:} Tools can have operation- and sequence-dependent configurations, which must be set up on a machine before processing operations.
  A configuration might imply a minimum/maximum number of operations to be performed before it can/must be changed. 
  Moreover, some operations require reconfiguring a machine, even if the same configuration was in place before.
  \item \textbf{Tool Family Specificities:} Certain functional areas of a fab have specific constraints. 
  For instance, steppers in the photolithography area have an \textit{lot-to-lens} dedication, enforcing the product to revisit the same tool in a future step of the route.
  Metrology steps performing quality control can be skipped by a production process with some probability.
  Dry etching steps have a Critical Queue Time (CQT) constraint enforcing the next operation to be performed with no more than a given amount of time delay.
  \item \textbf{Lot Priorities:} Lots can have different priorities: \textit{Super Hot Lots} have the highest and \textit{Hot Lots} have a higher priority at all stages of processing than regular lots. 
  \item \textbf{Variable Lot Size:} The processing time of certain operations might depend on the number of wafers in a lot, which can vary from one lot to another.
\end{enumerate}
\let\thefootnote\relax\footnotetext{Code and dataset available at: \url{https://github.com/ingambe/RL4SemiconductorFabSched}}

\textit{SMT2020} provides hierarchical dispatching heuristics, combining $4$ stages in the order of significance as follows:

\begin{enumerate}[left=0pt]
  \item \textbf{Critical Queue Time:} Lots with a CQT constraint in the previous operation have the top priority.
  \item \textbf{Lot Priorities:} Lots must be scheduled according to their priorities, i.e, \textit{Super Hot Lots} scheduled first, then \textit{Hot Lots}, and finally the regular lots.
  \item \textbf{Setup Avoidance:} Reconfiguration of a tool is time-consuming, and therefore, it is preferable to schedule operations in a way that avoids switching the setup.
  \item \textbf{Generic Dispatching Heuristic:} To break ties, the \textit{HV/LM} and \textit{LV/HM} scenarios use First-Come First-Serve (FCFS) or Critical Ratio (CR) heuristics, respectively, where  CR considers the due date and the remaining processing time required to complete a lot.
\end{enumerate}

The resulting simulation models are substantially different from the static and deterministic JSP, where all job and machine parameters are known in advance and the environment is not changing during the computation and execution of a schedule.
In contrast, the \textit{SMT2020} scenarios are dynamic and stochastic as changes in the environment, like the release of new lots with different priorities or unscheduled downtimes of machines, might occur with some probability at any point in time.

\section{Related Work} \label{sec:background}
The Semiconductor Factory Scheduling Problem (SFSP) is often decomposed into tool scheduling problems (single machine job-shop), work center scheduling (parallel machine job-shop), or work area scheduling (flexible job-shop) \citep{monchSurveyProblemsSolution2011}.
However, decomposition approaches merely approximate the SFSP since combinatorial optimization methods, like \emph{Constraint Programming} (CP) or \emph{Mixed Integer Programming} (MIP), do not scale up to the full problem due to the high implementation efforts and long computation times \citep{brankeAutomatedDesignProduction2016}. 
That is, a realistic instance of SFSP contains millions of operations to schedule each year, thus going far beyond the capabilities of modern optimization algorithms \cite{ColT19CPeval,KovacsCPKostwein}.
On the other hand, deterministic approximations of SFSP might reduce the robustness of schedules in presence of unexpected events, like machine breakdowns or reworks. Finally, SFSP is a dynamic problem where new customer orders constantly arrive, requiring to recompute a schedule once the existing one gets inconsistent with new data.

Combined methods try to mitigate scalability issues by applying combinatorial optimization only for bottleneck machines (often lithographic tools), while using dispatching heuristics for other operations \citep{govindOperationsManagementAutomated2008,hamPracticalTwoPhaseApproach2015,klemmtSimulationbasedOptimizationVs2009}.
Often, such partial optimization approaches only marginally improve over 
deterministic heuristic solutions,
which have in turn been shown to be less effective than meta-heuristics constructed for stochastic SFSP-like settings \citep{nguyenComputationalStudyRepresentations2013}.

The drawbacks of combinatorial optimization methods on large and stochastic scheduling problems made dispatching heuristics a de-facto industry standard.
Respective scheduling systems only manage to approximate optimal solutions, but they compute results in seconds and, therefore, can react to changes in real time \citep{bauernhanslProductionSchedulingComplex,hamPracticalTwoPhaseApproach2015}.
Any dispatching heuristic is a handcrafted heuristic designed by experts taking experimental studies and personal experience on typical scheduling scenarios into account.
The quality of different dispatching heuristics is evaluated using SFSP simulators with various objectives like reducing mean cycle-time, average tardiness, and delayed lots.
Although algorithms based on dispatching heuristics have shown good runtime results, there is no universal set of heuristics that work best for all evaluation scenarios. For instance, their performance depends on the current Work-In-Progress (WIP) \citep{nyhuisLogistischeKennlinienGrundlagen2012}, selected key performance indicators \citep{uzsoyPerformanceEvaluationDispatching1993}, and types of machines used \citep{fordyceMODELINGINTEGRATIONPLANNING2015}. 

Attempts to apply RL to semiconductor manufacturing aim to overcome the issues of dispatching heuristics in complex SFSP environments.
\cite{waschneckDeepReinforcementLearning2018} propose a multi-agent RL approach in which each agent represents a work center, i.e., a tool family.
This work models a factory with $4$ work centers and $48$ WIP lots to schedule, where transportation times and machine breakdowns are disregarded.
Agents are trained in two phases: \textit{(1)} only one agent (i.e., a work center) is trained while the others are controlled by a heuristic, and \textit{(2)} all agents controlling the $4$ work centers are trained together. 
Results obtained with the trained agents are comparable to the original dispatching strategy, but as the action space depends on the number of WIP lots, the approach
is not easy to generalize.
\citet{strickerReinforcementLearningAdaptive2018} propose a single-agent RL method for a testbed with $3$ work centers and a total of $8$ machines, without setup changes between different kinds of operations.
The dispatching agent can select from a set of $12$ actions to decide how to schedule a given lot.
Since the action space depends on the number of machines, a neural network trained on one fab cannot be reused for others and it might be difficult to apply this approach to thousands of machines as in \textit{SMT2020}. 
However, the trained agent improves the initial dispatching strategy significantly by $8\%$ better fab utilization. 


\section{Learning Algorithm}  \label{sec:learning}


This work aims at the development of a scheduler for the full SFSP, in which the objective is to efficiently dispatch the lots to schedule on the machines, so that metrics like mean cycle-time, average tardiness, and the number of delayed lots are minimized.
In this section, we formulate semiconductor fab dispatching decision making as an RL problem, present the agent network architecture, the self-supervision pre-training of the tool type encoding, and the agent training settings.

\subsection{Approach Overview} \label{subsec:approach}

Following the Markov Decision Process (MDP) formulation introduced by \cite{tasselReinforcementLearningDispatching2022}, the state-transition function of our deep RL agent considers sequences of lots rather than a fixed number of actions assigning lots to machines.
In particular, at any timepoint when a set of lots must be dispatched, the agent is trained to order lots in a sequence in which they are assigned to machines using a set-up avoidance scheme.
In the following, we formulate dispatching decisions on lot operations to be scheduled in terms of an MDP, including the state space~$\mathcal{S}$, action space~$\mathcal{A}$, reward~$r_t$ and transition $s_{t+1}\sim p(\;\cdot\mid s_t,a_t)$ functions.

\paragraph{State Space $\mathcal{S}$.} 
A state~$s_t \in \mathcal{S}$ is given by the set $\mathcal{L}^t$ of legal lots at timepoint~$t$. A lot $l$ is \emph{legal} if its operation sequence $O_l=(\tau_1,\dots,\tau_n)$ comprises an operation $\tau_m$ that can be scheduled to free resources at $t$. Moreover, each lot $l$ is associated with a set of features:
\begin{enumerate*}[label=\emph{(\arabic*)}]
    \item critical ratio, i.e., a ratio of the time remaining till the due date~$d_l$ to the expected processing time to finish the job; 
    \item remaining time until the deadline~$d_l$;
    \item total waiting time of the lot since its release;
    \item waiting time of the lot since the last operation;
    \item number of lot-to-lens dedications until lot completion, i.e., once a machine is used to perform a specific operation, it will always be used for remaining operations $\tau_i \in O_l$;
    \item lot priority $w_l$;
    \item sum of mean processing times for all remaining operations of the lot;
    \item minimum machine setup time to perform the current operation of the lot;
    \item mean processing time $p_\tau \in \mathbb{N}^+$ of the current operation $\tau \in O_l$;
    \item number of available machines with compatible setup;
    \item minimum batch size;
    \item maximum batch size; and 
    \item the family of the required tool group.
\end{enumerate*}

\paragraph{Action Space $\mathcal{A}$.} 
The action space~$\mathcal{A}$ is composed of all legal lots at timepoint~$t$, i.e., $\mathcal{A} = \mathcal{L}^t$. Hence, the action space is discrete, and the cardinality of~$\mathcal{A}$ may vary from one timepoint to another.
The agent receives as input a set of dispatchable lots for the current timepoint $t$, and provides as output the order in which to dispatch the given lots.
For each lot $l \in \mathcal{L}^t$, the network produces a scalar representing the priority of $l$. The ordered set of lots to dispatch is then obtained by sorting the lots according to their priority scores. 

\paragraph{Reward $r_t$.} 
Usually, SFSP uses the average cycle-time, tardiness, and the number of delayed lots for each lot type to compute the quality of a schedule. Unfortunately,  evaluations of these criteria provide meaningful results only when computed for a complete schedule and might be misleading for incomplete ones.
Therefore, we have designed an objective function assigning scores for any (in)complete schedule.
This function penalizes tardiness of a lot and long cycle-times while rewarding high throughputs. Moreover, the score for each lot is weighted based on its priority.

Our objective function comprises two parts: one for the lots already finished at the time of the evaluation and another for the lots still in the WIP.
Both parts are then summed together to obtain the final result.
Considering lots in the WIP for evaluating a schedule is particularly important in the case of a continuous scheduling environment such as SFSP, as one potential bias of a method is to only optimize metrics for the evaluated time windows but risk bringing the fab into a catastrophic state in the future.

The first part, only considering lots finished at the evaluation time $t$, is defined as:
\begin{empheq}[left={%
 O_{f} = 
 \displaystyle\sum_{k_i \in K}\;
 \frac{1}{\lvert k_i \rvert}
 \sum_{l \in k_i}\empheqlbrace}]{align*}
  w_l * (p + (c_l - d_l)) & \quad \text{if } c_l > d_l \nonumber\\
 \rlap{$0$} \phantom{w_l * (p + (c_l - d_l))} & \quad \text{otherwise}
\end{empheq}
where $K$ the set of all lot types, $\lvert k_i \rvert$ the number of lots of type $k_i$, $c_l$ the completion date and $d_l$ the deadline for lot $l$. 
The constant term $p$ allows to penalize delayed lots independently of whether the tardiness is large or small.
%
The second part, focusing on lots in the WIP at time $t$, is defined as:
\begin{equation*}
O_{p} = 
 \sum_{k_i \in K}\;
 \frac{1}{\lvert k_i \rvert}
 \sum_{l \in k_i} w_l * x_{i,l}
\end{equation*}
\begin{empheq}[left={x_{i,l}=\empheqlbrace}]{align*}
  p + (t - (d_l - (a_i * e_l))) & \quad\text{if } d_l - (a_i * e_l) < t \nonumber\\
   \rlap{$0$} \phantom{p+ (t - (d_l - (a_i * e_l)))}& \quad\text{otherwise}
\end{empheq}
where $a_i$ is the average cycle-time of lots of type $k_i$ computed on the finished lots, and $e_l$ the sum of mean left-to-process operations' processing time.
The term $d_l - (a_i * e_l)$ evaluates the deadline for lots in the WIP by forecasting the completion date of a given lot based on the average processing time for operations to be performed until completion.
Thus, the minimization of the objective function $O = O_{f} + O_{p}$ leads to the reduction of tardiness, increase in throughput, and decrease in the average cycle-time for each lot type, just as the original SFSP objective.
Note that, due to the specificities of our training approach, which employs a derivative-free optimization method, we do not require a continuous and differentiable reward function.

\begin{figure*}[ht]
  \centering
   \includegraphics[width=0.8\textwidth]{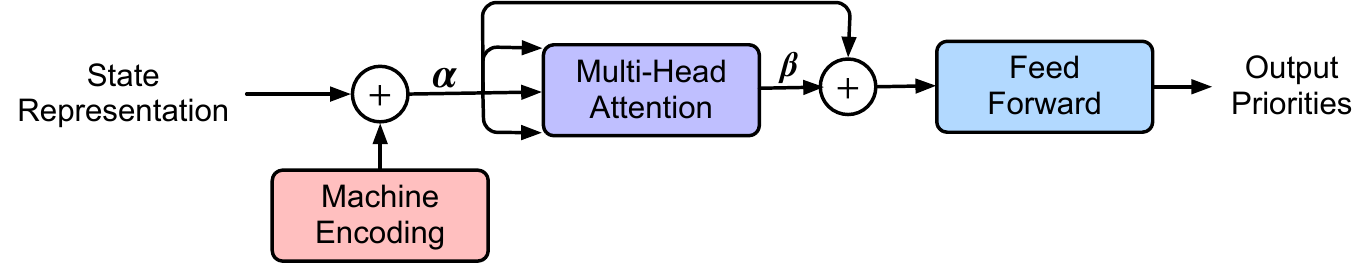}
   \caption{Policy network architecture with an embedding encoding, injecting information about the tool family required for an operation. The network outputs a priority score for each lot representation.}
   \label{Fig:architecture}
\end{figure*}

\paragraph{State Transition.}
At each decision timepoint $t$, the simulator provides information about the current state $s_t$ of a fab, and awaits an ordered list of lots to dispatch.
Similar to static dispatching heuristics, the ordered list is processed using a hierarchical strategy, first giving higher priority to lots with a CQT constraint, then lots' priorities (i.e., \emph{Hot Lots}), and finally the original order provided by the agent.   
The simulator then allocates the lots to their associated free machines in the given order until no free resources remain or all lots are scheduled.
Next, new events from the events stack are handled and the timepoint is incremented.
In \emph{SMT2020}, lots are assigned to machines using the following heuristic strategy:%
\begin{enumerate}[label=\emph{(\arabic*)},left=0pt]
 \item If there is a lot-to-lens dedication constraint, assign to the mandatory machine.
 \item If the current operation does not require a setup, assign to a machine without setup, if any. Otherwise, assign to the machine with the highest waiting time.
 \item If the operation requires a setup, assign to a machine with this setup, if available. Otherwise, assign to the machine with the lowest setup time.
\end{enumerate}
This strategy is greedy as it merely determines advantageous allocations at a timepoint $t$, does neither backtrack nor consider future incoming lots, i.e., no "lookahead". However, since \emph{SMT2020} supplies information about the legal lots at timepoint $t$ only, the machine allocation strategy is also used by our agent.

\subsection{Policy Architecture} \label{subsec:architecture}

The agent policy is a function $\pi_\theta(\mathcal{A}\mid s_t)$ that maps the current state of the simulator $s_t$ to the ordered set $\mathcal{A}$ of legal actions (lots) to dispatch for the current timepoint $t$.
$\pi_\theta$ is implemented as a deep neural network of the structure illustrated in Figure~\ref{Fig:architecture}.

Before feeding the state representation to the policy neural network, we rescale the features of the legal lots using a virtual batch normalization from statistics over a $2$-months horizon collected before training.
For each lot representation $l \in s_t$, we inject information about the tool family required to perform the current operation
, using a learned encoding, and keep the first $12$ features as the lot representation (i.e., the $13$-th feature is the tool family index that is only used by the learned encoding).
Encodings are often used to inject positional information via absolute positional embeddings \citep{vaswaniAttentionAllYou2017} or relative bias factors \citep{shawSelfAttentionRelativePosition2018},
where a positional encoding is either fixed or learned \citep{havivTransformerLanguageModels2022}.
In our case, the injected information is the family of the tool required to perform the current operation on a lot.
We learn an associate machine encoding for each of the $12$ tool families and add it to the input lot representation.
Moreover,
we experimented with input embedding before the machine encoding, similar to \citep{dosovitskiyImageWorth16x162021,vaswaniAttentionAllYou2017}, but observed degradation of performance.

The transformed state representation $s'_t$ is passed to a double-head self-attention mechanism and a simple, position-wise, fully connected Feed-Forward Network (FFN).
We employ a residual connection with scaling factor \citep{heDeepResidualLearning2015,szegedyInceptionv4InceptionResNetImpact} around $s'_t$ and the FFN:%
$$\boldsymbol{y} = \alpha \boldsymbol{x} + \beta f(\alpha \boldsymbol{x})$$
with $\alpha$ and $\beta$ as two learnable parameters.
The multi-head attention layer allows the network to determine the priority of each lot not only based on the features of the given lot but taking all lots to schedule at the current timepoint into account.
It contains two attention heads, each of dimension $d_{\mathit{head}} = 6$.
The FFN comprises three linear transformations of dimension $16$ with a \emph{tanh} activation in between:
\begin{equation*}
   \mathrm{FFN}(x)=\tanh(\tanh(xW_1 + b_1) W_2 + b_2) W_3 + b_3
\end{equation*} 
Similar to the transformer architecture \citep{vaswaniAttentionAllYou2017}, we apply the FFN to each lot representation separately. 
The inner layers have dimensionality $d_{\mathit{FFN}}=16$, while the network outputs a scalar representing the estimated lot priority.

\subsection{Training} \label{subsec:training}

With an average of $7\,254\,125$ allocated steps per year, training the policy network is challenging since the search space 
is highly combinatorial.
Also, agents need to be trained with a long horizon since some products have a long cycle-time with completion up to three months in the future, while disruptions such as machine breakdown occur relatively rarely.
Therefore, we use a training horizon of $6$ months, as it provides a good balance between training runtime and the diversity of events handled by an agent.

\subsubsection{Self-Supervised Learning.} \label{subsec:self_supervised}

\begin{figure*}[ht]
  \centering
   \includegraphics[width=\textwidth]{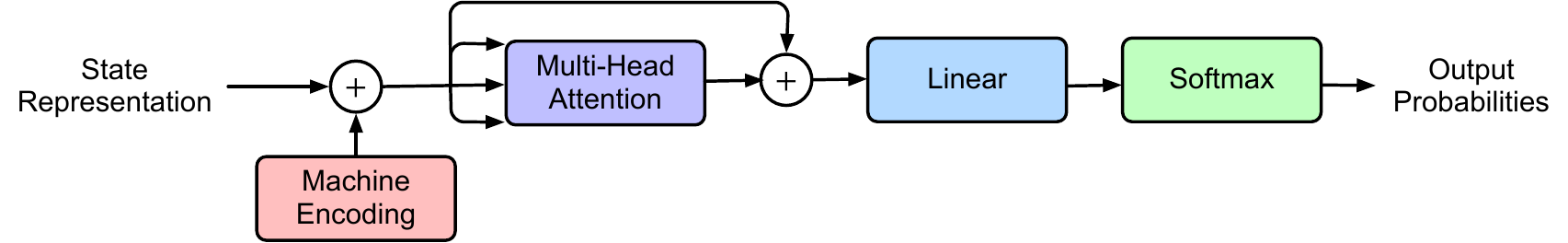}
   \caption{Self-supervised pretext task architecture. The lot representation is extended with information about the required tool family, which then must be correctly predicted by the network.}\label{Fig:self_supervised}
\end{figure*}

While the multi-head attention and FFN layers are trained together, the machine encoding is trained separately using a self-supervised pretext task \citep{doerschUnsupervisedVisualRepresentation2016, gidarisUnsupervisedRepresentationLearning2018}.
The goal of the machine encoding is to inject information about the required tool family into the lot representation.
This information is very important as different tool families have very different processes and constraints, for example, lithographic tools are usually considered critical due to their scarcity, etching steps often have a coupling constraint with their next operation, and other tool families also have their specificities.
To achieve this goal, we propose to train a network as shown in Figure~\ref{Fig:self_supervised}, having a similar architecture as the policy network. 
The network receives a set of lot representations as input and yields a probability distribution over all possible tool families for each lot as output.

Therefore, given a set of $N$ training state representations $D = \{x_i\}_{i=0}^{N}$ sampled from a $2$-months horizon simulation, the self-supervised training objective that the model must learn to solve is:
\begin{equation*}
  \min_{\theta} \frac{1}{N}\sum_{i=1}^{N} loss(x_i,\theta) + \lambda \| \theta_{e} \|^2
\end{equation*}
where the loss function $loss(.)$ is the cross-entropy between the predicated and the ground-truth tool family, $\theta$ the parameters of the network, $\theta_{e}$ the parameters of the machine encoding, and $\lambda$ the regularization constant.
This $L_2$ regularization is required to prevent the encoding from taking too large values and, thus, destroying the information about lot features.
The $\lambda=0.2$ regularization constant is tuned such that the performance on the downstream task (i.e., the fab dispatching) is maximized.

The pretext task is trained until convergence using Stochastic Gradient Descent (SGD) and backpropagation.
The multi-head attention and the linear layer are then discarded, and only the learned machine encoding weights are kept frozen for the downstream task.

\subsubsection{Policy Network Training.} \label{subsec:es}

Objective functions of scheduling problems and discrete sets of possible job allocations result in learning environments with non-smooth objectives, which are unsuitable for training by backpropagating gradients. 
We overcome this issue by using a combination of two methods: estimation of a gradient with the Natural Evolution Strategies (NES) algorithm \cite{salimansEvolutionStrategiesScalable2017} and a modern gradient-based optimization algorithm to update weights of the policy network. 

NES belongs to a broad family of Evolution Strategies (ES), which are black-box, derivative-free methods inspired by natural evolution. 
When applied to the training of a policy network $\pi_\theta$, NES methods use the parameter vector $\theta$ of the policy to generate a population of parameter vectors $\theta'_i \sim \mathcal{N}(\theta, \sigma^2 I)$ by adding noise sampled from an isotropic multivariate Gaussian to the original parameter vector $\theta$, thus obtaining a set of $n$ successor networks. 
Then, each of the $n$ networks in the population is used to evaluate an objective function, the episodic reward function in our case, to estimate the \emph{natural gradient} for the policy network:
\begin{equation*}
    \nabla_\theta F(\pi_\theta) \approx \frac{1}{\sigma_t * N} \sum_{i=1}^N F(\pi_{\theta_i^\prime})\epsilon_i
\end{equation*}
where $\epsilon_i \sim \mathcal{N}(0,\mathbb{I})$ is sampled from a multivariate Gaussian, $\theta_i^\prime = \theta + \sigma_t \epsilon_i$, $F(\pi_{\theta_i^\prime})$ is estimated using a single trajectory, and $\sigma_t > 0$ is a constant defined as $\sigma_t = 0.975^{t}\sigma$ with $\sigma=0.005$.


The estimated natural gradient is provided to Adam optimizer \citep{kingmaAdamMethodStochastic2017} with $\beta_1=0.9$, $\beta_2=0.98$, and $\epsilon=10^{-4}$, which updates the parameters $\theta$ of the policy network.
Within the $i$-th run, we decay the learning rate with a cosine annealing \citep{loshchilovSGDRStochasticGradient2017} 
as follows:%
\begin{equation*}
	\eta_i = \frac{\eta_{max}}{2} \left(1 + \cos\left(\frac{i}{i_{max}} \pi\right)\right)
\end{equation*}
where $\eta_{max} = 0.01$ is the maximum learning rate, and $i_{max}=40$ the maximum number of iterations.

The training algorithm continues to repeat these steps till convergence.
Because NES works with 
function evaluations, there is no need to rely on a continuous reward function.
Therefore, we can use the previously defined unbiased and episodic objective function.

\begin{table*}[ht]
  \begin{subtable}[h]{0.97\textwidth}
    \centering

  \resizebox{\textwidth}{!}{
  \begin{tabular}{l l r@{$\displaystyle \,\pm\,$}l r@{$\displaystyle \,\pm\,$}l r@{$\displaystyle \,\pm\,$}l}
  \toprule
  \multicolumn{2}{l}{\textbf{Lot Type}}      & \multicolumn{2}{c}{\textbf{Ours}}                & \multicolumn{2}{c}{\textbf{Hierarchical CR}}              & \multicolumn{2}{c}{\textbf{Hierarchical FIFO}} \\
  \midrule 
\textbf{Super Hot Lots} &   On-Time              & \valstdb{42.50\%}{13.94}   & \valstd{41.84\%}{12.45}  & \valstd{38.06\%}{10.64}  \\
\small{Theoretical Cycle-Time: $\mathbf{24.75}$} &    Cycle-Time              & \valstdb{33.61}{0.51}   & \valstd{33.89}{0.43}  & \valstd{33.99}{0.44}  \\ 
\addlinespace 
\textbf{Hot Lots}                &    On-Time              & \valstdb{55.86\%}{3.58}   & \valstd{46.76\%}{3.77}  & \valstd{47.66\%}{3.74}    \\
\small{Theoretical Cycle-Time: 
$\mathbf{19.65}$}  &    Cycle-Time       & \valstdb{26.54}{0.14}   & \valstd{26.94}{0.13}  & \valstd{26.91}{0.12}            \\ 
\addlinespace 
\textbf{Regular Lots} &   On-Time              & \valstdb{44.91\%}{2.36}   & \valstd{0.00\%}{0.00}  & \valstd{0.00\%}{0.00}  \\
\small{Theoretical Cycle-Time: $\mathbf{17.31}$} &    Cycle-Time       & \valstdb{52.24}{1.80}   & \valstd{70.80}{0.60}  & \valstd{72.57}{0.61}  \\ 
\midrule 
& \textbf{Cost}    & \valstdb{33\,171.37}{1\,655.53}                  & \valstd{39\,813.37}{899.89}              & \valstd{38\,013.67}{1\,327.73}        \\
  \bottomrule
  \end{tabular}
  }
  \caption{High-Volume/Low-Mix (\textit{HV/LM})}\label{tbl:HVLM}

\end{subtable}
\vspace{5pt}

\begin{subtable}[h]{0.97\textwidth}

  \centering
  \resizebox{\textwidth}{!}{
  \begin{tabular}{l l r@{$\displaystyle \,\pm\,$}l r@{$\displaystyle \,\pm\,$}l r@{$\displaystyle \,\pm\,$}l}
  \toprule
  \multicolumn{2}{l}{\textbf{Lot Type}}      & \multicolumn{2}{c}{\textbf{Ours}}                & \multicolumn{2}{c}{\textbf{Hierarchical CR}}              & \multicolumn{2}{c}{\textbf{Hierarchical FIFO}} \\
  \midrule 
\textbf{Super Hot Lots} &   On-Time              & \valstdb{97.38\%}{4.63}   & \valstd{97.19\%}{4.72}  & \valstd{95.04\%}{5.44}  \\
\small{Theoretical Cycle-Time: $\mathbf{24.75}$} &    Cycle-Time              & \valstdb{33.64}{0.44}   & \valstd{33.73}{0.42}  & \valstd{33.86}{0.45}  \\ 
\addlinespace 
\textbf{Hot Lots} &   On-Time              & \valstd{96.03\%}{3.20}   & \valstdb{96.34\%}{3.04}  & \valstd{95.38\%}{3.35}  \\
\small{Theoretical Cycle-Time: $\mathbf{17.31}$} &    Cycle-Time              & \valstdb{23.67}{0.23}   & \valstd{23.71}{0.23}  & \valstd{23.80}{0.23}  \\ 
\addlinespace 
 \textbf{Regular Lots} &   On-Time              & \valstdb{56.47\%}{2.36}   & \valstd{0.00\%}{0.00}  & \valstd{0.06\%}{0.09}  \\
 \small{Theoretical Cycle-Time: $\mathbf{17.31}$} &    Cycle-Time       & \valstdb{44.15}{1.80}   & \valstd{52.25}{0.30}  & \valstd{52.38}{0.63}  \\ 
\midrule 
& \textbf{Cost}    & \valstdb{94\,349.38}{2\,716.51}                  & \valstd{108\,855.67}{2\,044.28}              & \valstd{102\,170.71}{2\,497.89}        \\
  \bottomrule
  \end{tabular}
  }
  \caption{Low-Volume/High-Mix (\textit{LV/HM})}\label{tbl:LVHM}

\end{subtable}
  \caption{
  Comparison against the original dispatching heuristics on the \textit{SMT2020} testbed for the \textit{HV/LM} and \textit{LV/HM} scenarios with $6$ months of simulation.
  For each lot priority level, we report the aggregated mean and standard deviation of the percentage of lots on-time and the cycle-time (see the appendix for more detailed results).
  Our approach outperforms the original ones in both scenarios, especially for regular lots, which represent the major part of all lots in production.
  }
  \label{tbl:LVHM/HVLM}
\end{table*}

\section{Empirical Evaluation} \label{sec:eval}

We evaluated our approach against the original \emph{SMT2020} hierarchical dispatching heuristics described in the background, 
using a machine equipped with AMD EPYC $7452$ CPU and $128$GB of RAM for
benchmarking and training, with the training iterations taking $3$ hours on average.
As discussed in the related work, to our knowledge, there are no CP- or MIP-based approaches in the literature for handling such large problems of stochastic, dynamic, and continuous nature. Therefore, as a baseline, we rely on proposed heuristic methods, which are the de-facto standard in semiconductor industry \cite{govindOperationsManagementAutomated2008,hamPracticalTwoPhaseApproach2015}.

For each dataset, we run all dispatching strategies with a horizon of $6$ months with $100$ different random seeds and report the percentage of lots that meet the deadlines defined at release time along with the cycle-time, i.e., the average number of days from release to completion per given lot type.
We provide the mean and the standard deviation of each metric collected for agents trained in a considered scenario.
The penalty constant was set to $p=10$ encouraging the learner to prefer fewer delayed lots with a larger delay over a larger number of delayed lots with a very short delay each. 
In comparison, experimentation conducted with $p=5$ leads to fewer lots on time but a shorter average cycle-time.

%
%
Our evaluation settings are realistic since the production scenario rarely changes during the lifetime of a fab.
Due to page constraints, Table \ref{tbl:LVHM/HVLM} only reports the aggregated metrics for each lot priority level.
To get a full picture of the performance of our approach, we invite the reader to refer to the appendix, providing more fine-grained metrics per lot type.

As we can see in Table \ref{tbl:HVLM}, for the \textit{HV/LM} dataset, our approach outperforms the state of the art, obtaining solutions of $16.68\%$ lower cost when compared to the best-performing dispatching heuristic.
Both the average tardiness and the cycle-time of lots are reduced.
While the absolute reduction in solution cost is already significant, as the unknown ``optimal'' solution (i.e., the optimal solution for each seed) has a cost greater than $0$, we have an even higher relative improvement.
Focusing on \textit{Regular Lots}, the vast majority of all lots, the original dispatching heuristics are not capable of scheduling them to meet the deadline. In contrast, our approach meets the deadline for $44.91\%$ of the lots and reduces the cycle-time by $26.21\%$ when compared to the \textit{CR} dispatching heuristic, thus substantially decreasing the average time from release to completion of a lot.

Aggregated results for the \textit{LV/HM} dataset are reported in Table \ref{tbl:LVHM}.
As with the \textit{HV/LM} scenario, our method outperforms the state of the art, obtaining solutions of $7.66\%$ lower cost when compared to the best-performing dispatching heuristic. 
Although the metrics obtained for \textit{Hot Lots} are very similar, our approach is always capable of reducing the cycle-time and achieving overall lower tardiness of lots. 
Also paralleling the \textit{HV/LM} scenario, the agent vastly outperforms static dispatching heuristics when focusing on \textit{Regular Lots}, meeting the deadline for $56.47\%$ of them. In comparison, \textit{Regular Lots} scheduled by standard heuristics met their deadlines in almost none of the cases. Furthermore, our approach greatly reduces tardiness and the cycle-time by about $15.5\%$, thus making better usage of fab resources.

For extended evaluations, including results per lot type, comparison with additional dispatching heuristics from the literature, and a $1$-year evaluation horizon, we invite the reader to refer to the appendix.
In all of these scenarios, we observe similar results as described above. 
Our approach outperforms other methods from the literature, improving almost all metrics while only adding a negligible overhead on computation time with an, on average, $1.02$ ms longer decision time per timepoint.

\section{Conclusion} \label{sec:conclusion}

We present a new scheduling method capable of improving the usage of semiconductor fab resources.
Unlike prior works focusing on toy datasets with a handful of products, machines, or constraints, we aim at realistic semiconductor manufacturing models in full complexity and scale.
Using a combination of self-supervised learning to learn an embedding encoding information about available tool families, and reinforcement learning to train an attention-based model capable of considering all lots to allocate, we trained an agent to dispatch lots to machines efficiently. 
Extensive evaluation against a wide range of dispatching heuristics in the literature, including those proposed for the considered datasets, shows the ability of our approach to substantially improve semiconductor scheduling capacities and achieve better usage of a fab's resources.

While these initial results are encouraging, many challenges remain.
Mainly, applying our approach in production first requires building trust of operators responsible for the production.
In fact, it can be challenging to convince operators to follow decisions made by such a black-box system, especially if some decisions might initially seem counterintuitive.
Another challenge is to provide a more global vision of the fab to the agent by including lots currently processed by a tool in the state representation.
It would allow the agent to anticipate which lots are getting ready next and possibly decide to leave a tool unallocated for a soon-to-be-available lot. 
In this regard, with a WIP exceeding $3\,000$ lots, the squared complexity of the dot-product attention mechanism used in our current architecture might, however, limit the possibilities to incorporate currently processed lots.

\bibliography{main}

\newpage
\appendix



\begin{table*}[h!]
  \begin{subtable}[h]{\textwidth}
    \centering

  \resizebox{0.8\textwidth}{!}{
  \begin{tabular}{l l r@{$\displaystyle \,\pm\,$}l r@{$\displaystyle \,\pm\,$}l r@{$\displaystyle \,\pm\,$}l}
  \toprule
  \multicolumn{2}{c}{\textbf{Lot Type}}       & \multicolumn{2}{c}{\textbf{Ours}}                & \multicolumn{2}{c}{\textbf{Hierarchical CR}}              & \multicolumn{2}{c}{\textbf{Hierarchical FIFO}} \\
  \midrule 
\textbf{Super Hot Lots 3} &   On-Time              & \valstdb{42.50\%}{13.94}   & \valstd{41.84\%}{12.45}  & \valstd{38.06\%}{10.64}  \\
\small{Theoretical Cycle-Time: $\mathbf{24.75}$} &    Cycle-Time              & \valstdb{33.61}{0.51}   & \valstd{33.89}{0.43}  & \valstd{33.99}{0.44}  \\ 
\addlinespace 
\textbf{Hot Lots 3}                &    On-Time              & \valstdb{51.77\%}{3.71}   & \valstd{42.99\%}{3.58}  & \valstd{43.09\%}{3.85}    \\
\small{Theoretical Cycle-Time: $\mathbf{24.75}$}              &    Cycle-Time              & \valstdb{33.56}{0.17}   & \valstd{34.02}{0.14}  & \valstd{34.02}{0.13}            \\ 
\textbf{Hot Lots 4}  &    On-Time              & \valstdb{59.94\%}{3.45}   & \valstd{50.53\%}{3.95}  & \valstd{52.26\%}{3.63}   \\
\small{Theoretical Cycle-Time: $\mathbf{14.54}$}  &    Cycle-Time              & \valstdb{19.52}{0.11}   & \valstd{19.86}{0.12}  & \valstd{19.80}{0.10}             \\ 
\addlinespace 
\textbf{Regular Lots 3} &    On-Time              & \valstdb{51.72\%}{2.47}   & \valstd{0\%}{0}  & \valstd{0\%}{0}          \\
\small{Theoretical Cycle-Time: $\mathbf{24.75}$} &    Cycle-Time              & \valstdb{63.71}{1.90}   & \valstd{90.77}{0.53}  & \valstd{93.61}{0.71}            \\ 
\textbf{Regular Lots 4} &    On-Time              & \valstdb{38.1\%}{2.04}   & \valstd{0\%}{0}  & \valstd{0\%}{0}        \\
\small{Theoretical Cycle-Time: $\mathbf{14.54}$} &    Cycle-Time             & \valstdb{40.77}{1.7}   & \valstd{50.83}{0.66}  & \valstd{51.52}{0.5}            \\ 
\midrule 
& \textbf{Cost}    & \valstdb{33\,171.37}{1\,655.53}                  & \valstd{39\,813.37}{899.89}              & \valstd{38\,013.67}{1\,327.73}        \\
  \bottomrule
  \end{tabular}
  }

  \caption{High-Volume/Low-Mix (\textit{HV/LM})}\label{tbl:HVLM_full}

\end{subtable}

\begin{subtable}[h]{\textwidth}
  \centering

\resizebox{0.8\textwidth}{!}{
  \begin{tabular}{l l r@{$\displaystyle \,\pm\,$}l r@{$\displaystyle \,\pm\,$}l r@{$\displaystyle \,\pm\,$}l}
  \toprule
  \multicolumn{2}{c}{\textbf{Lot Type}}       & \multicolumn{2}{c}{\textbf{Ours}}                & \multicolumn{2}{c}{\textbf{Hierarchical CR}}              & \multicolumn{2}{c}{\textbf{Hierarchical FIFO}} \\
  \midrule 
\textbf{Super Hot Lots 3} &   On-Time              & \valstdb{97.38\%}{4.63}   & \valstd{97.19\%}{4.72}  & \valstd{95.04\%}{5.44}  \\
\small{Theoretical Cycle-Time: $\mathbf{24.75}$} &    Cycle-Time              & \valstdb{33.64}{0.44}   & \valstd{33.73}{0.42}  & \valstd{33.86}{0.45}  \\ 
\addlinespace 
\textbf{Hot Lots 1} &   On-Time              & \valstd{97.08\%}{3.29}   & \valstdb{97.60\%}{2.77}  & \valstd{96.4\%}{3.42}  \\
\small{Theoretical Cycle-Time: $\mathbf{21.75}$} &    Cycle-Time              & \valstdb{29.81}{0.29}   & \valstd{29.9}{0.27}  & \valstd{29.97}{0.26}  \\ 
\textbf{Hot Lots 2} &   On-Time              & \valstd{98.65\%}{2.2}   & \valstdb{99.23\%}{1.8}  & \valstd{98.15\%}{2.59}  \\
\small{Theoretical Cycle-Time: $\mathbf{23.3}$} &    Cycle-Time              & \valstdb{31.5}{0.24}   & \valstd{31.66}{0.26}  & \valstd{31.81}{0.25}  \\ 
\textbf{Hot Lots 3} &   On-Time              & \valstdb{97.93\%}{2.55}   & \valstd{97.90\%}{2.87}  & \valstd{96.93\%}{3.38}  \\
\small{Theoretical Cycle-Time: $\mathbf{24.75}$} & Cycle-Time  & \valstdb{33.56}{0.32}   & \valstd{33.71}{0.27}  & \valstd{33.88}{0.3}  \\ 
\textbf{Hot Lots 4} &   On-Time              & \valstd{96.62\%}{3.55}   & \valstdb{96.92\%}{3.42}  & \valstd{95.91\%}{3.67}  \\
\small{Theoretical Cycle-Time: $\mathbf{14.54}$} & Cycle-Time  & \valstdb{19.87}{0.22}   & \valstd{19.94}{0.22}  & \valstd{20}{0.21}  \\ 
\textbf{Hot Lots 5} &   On-Time              & \valstdb{94.17\%}{3.03}   & \valstd{94.05\%}{3.06}  & \valstd{94.20\%}{2.88}  \\
\small{Theoretical Cycle-Time: $\mathbf{10.10}$} & Cycle-Time  & \valstdb{14.09}{0.17}   & \valstd{14.11}{0.17}  & \valstd{14.06}{0.17}  \\ 
\textbf{Hot Lots 6} &   On-Time              & \valstd{96.42\%}{2.62}   & \valstdb{97.15\%}{2.47}  & \valstd{96.64\%}{2.48}  \\
\small{Theoretical Cycle-Time: $\mathbf{12.95}$} & Cycle-Time  & \valstdb{17.47}{0.16}   & \valstd{17.53}{0.21}  & \valstd{17.53}{0.2}  \\ 
\textbf{Hot Lots 7} &   On-Time              & \valstdb{95.22\%}{3.68}   & \valstd{94.82\%}{3.54}  & \valstd{94.28\%}{3.87}  \\
\small{Theoretical Cycle-Time: $\mathbf{15.46}$} & Cycle-Time  & \valstdb{21.02}{0.21}   & \valstd{21.05}{0.22}  & \valstd{21.11}{0.23}  \\ 
\textbf{Hot Lots 8} &   On-Time              & \valstdb{92.41\%}{3.71}   & \valstd{92.39\%}{3.77}  & \valstd{90.32\%}{4.01}  \\
\small{Theoretical Cycle-Time: $\mathbf{16.02}$} & Cycle-Time  & \valstdb{22.21}{0.21}   & \valstd{22.21}{0.22}  & \valstd{22.33}{0.24}  \\ 
\textbf{Hot Lots 9} &   On-Time              & \valstd{95.96\%}{3.28}   & \valstdb{96.27\%}{3.37}  & \valstd{95.03\%}{3.54}  \\
\small{Theoretical Cycle-Time: $\mathbf{16.95}$} & Cycle-Time  & \valstd{23.16}{0.2}   & \valstdb{23.05}{0.21}  & \valstd{23.22}{0.23}  \\ 
\textbf{Hot Lots 10} &   On-Time              & \valstd{95.88\%}{4.1}   & \valstdb{97.04\%}{3.33}  & \valstd{95.96\%}{3.64}  \\
\small{Theoretical Cycle-Time: $\mathbf{17.32}$} & Cycle-Time  & \valstd{24.01}{0.28}   & \valstdb{23.96}{0.25}  & \valstd{24.08}{0.25}  \\ 
\addlinespace 
 \textbf{Regular Lots 1} &   On-Time              & \valstdb{65.80\%}{1.77}   & \valstd{0\%}{0}  & \valstd{0\%}{0}  \\
 \small{Theoretical Cycle-Time: $\mathbf{21.75}$} &    Cycle-Time       & \valstdb{50.77}{1.67}   & \valstd{61.76}{0.33}  & \valstd{69.76}{0.78}  \\ 
 \textbf{Regular Lots 2} &   On-Time              & \valstdb{69.58\%}{1.8}   & \valstd{0\%}{0}  & \valstd{0\%}{0}  \\
 \small{Theoretical Cycle-Time: $\mathbf{23.3}$} &    Cycle-Time    & \valstdb{49.31}{1.26}   & \valstd{70.22}{0.45}  & \valstd{70.84}{0.76}  \\ 
 \textbf{Regular Lots 3} &   On-Time              & \valstdb{58.29\%}{2.42}   & \valstd{0\%}{0}  & \valstd{0\%}{0}  \\
 \small{Theoretical Cycle-Time: $\mathbf{24.75}$} & Cycle-Time  & \valstdb{61.85}{2.38}   & \valstd{74.40}{0.46}  & \valstd{78.75}{0.88}  \\ 
 \textbf{Regular Lots 4} &   On-Time              & \valstdb{46.02\%}{3.7}   & \valstd{0\%}{0}  & \valstd{0\%}{0}  \\
 \small{Theoretical Cycle-Time: $\mathbf{14.54}$} & Cycle-Time  & \valstdb{40.85}{2.58}   & \valstd{44.7}{0.34}  & \valstd{44.88}{0.49}  \\ 
 \textbf{Regular Lots 5} &   On-Time              & \valstdb{24.3\%}{2.12}   & \valstd{0\%}{0}  & \valstd{0\%}{0}  \\
 \small{Theoretical Cycle-Time: $\mathbf{10.10}$} & Cycle-Time  & \valstd{41.57}{2.48}   & \valstdb{31.28}{0.19}  & \valstd{33.16}{0.53}  \\ 
 \textbf{Regular Lots 6} &   On-Time              & \valstdb{58.18\%}{2.70}   & \valstd{0\%}{0}  & \valstd{0.59\%}{0.92}  \\
 \small{Theoretical Cycle-Time: $\mathbf{12.95}$} & Cycle-Time  & \valstdb{33.55}{1.57}   & \valstd{43.54}{0.35}  & \valstd{39.03}{0.59}  \\ 
 \textbf{Regular Lots 7} &   On-Time              & \valstdb{65.29\%}{2.24}   & \valstd{0\%}{0}  & \valstd{0\%}{0}  \\
 \small{Theoretical Cycle-Time: $\mathbf{15.46}$} & Cycle-Time  & \valstdb{36.7}{1.42}   & \valstd{45.08}{0.31}  & \valstd{47.73}{0.52}  \\ 
 \textbf{Regular Lots 8} &   On-Time              & \valstdb{67.04\%}{1.83}   & \valstd{0\%}{0}  & \valstd{0\%}{0}  \\
 \small{Theoretical Cycle-Time: $\mathbf{16.02}$} & Cycle-Time  & \valstdb{38.43}{1.50}   & \valstd{52.13}{0.36}  & \valstd{51.81}{0.65}  \\ 
 \textbf{Regular Lots 9} &   On-Time              & \valstdb{53.25\%}{2.66}   & \valstd{0\%}{0}  & \valstd{0\%}{0}  \\
 \small{Theoretical Cycle-Time: $\mathbf{16.95}$} & Cycle-Time  & \valstdb{44.48}{1.60}   & \valstd{48.24}{0.25}  & \valstd{33.86}{0.45}  \\ 
 \textbf{Regular Lots 10} &   On-Time              & \valstdb{56.97\%}{2.39}   & \valstd{0\%}{0}  & \valstd{0\%}{0}  \\
 \small{Theoretical Cycle-Time: $\mathbf{17.32}$} & Cycle-Time  & \valstdb{43.95}{1.67}   & \valstd{51.13}{0.29}  & \valstd{54.01}{0.63}  \\ 
\midrule 
& \textbf{Cost}    & \valstdb{94\,349.38}{2\,716.51}                  & \valstd{108\,855.67}{2\,044.28}              & \valstd{102\,170.71}{2\,497.89}        \\
  \bottomrule

  \end{tabular}
  }
  \caption{Low-Volume/High-Mix (\textit{LV/HM})}\label{tbl:LVHM_full}

\end{subtable}
\caption{
  Detailed results of the comparison against the original dispatching rules on the \textit{SMT2020} testbed for the \textit{HV/LM} and \textit{LV/HM} scenarios with $6$ months of simulation, providing aggregated metrics for each lot priority level.
  For each lot type, we report the mean and standard deviation of the percentage of lots on-time and the cycle-time.
  Our approach outperforms the original one on both datasets.}\label{tbl:HVLM_LVHM_full}
\end{table*} \label{section:LVHM/HVLM_full}



\begin{table*}[ht]
    \begin{subtable}[h]{\textwidth}
      \centering
  
    \resizebox{\textwidth}{!}{
  \begin{tabular}{l l r@{$\displaystyle \,\pm\,$}l r@{$\displaystyle \,\pm\,$}l r@{$\displaystyle \,\pm\,$}l r@{$\displaystyle \,\pm\,$}l r@{$\displaystyle \,\pm\,$}l}
  \toprule
  \multicolumn{2}{c}{\textbf{Lot Type}}      & \multicolumn{2}{c}{\textbf{Ours} }      &          \multicolumn{2}{c}{\textbf{Hierarchical SPT}} & \multicolumn{2}{c}{\textbf{Hierarchical SRPT}} & \multicolumn{2}{c}{\textbf{Hierarchical EDD}} & \multicolumn{2}{c}{\textbf{Hierarchical LS}} \\
  \midrule 
\textbf{Super Hot Lots 3} &   On-Time              & \valstd{42.50\%}{13.94}    & \valstd{41.76\%}{11.48} & \valstd{37.82\%}{12.76} & \valstd{37.08\%}{15.20} & \valstdb{43.37\%}{13.01} \\
\small{Theoretical Cycle-Time: $\mathbf{24.75}$} &    Cycle-Time              & \valstdb{33.61}{0.51}    & \valstd{33.81}{0.45} & \valstd{33.95}{0.42} & \valstd{33.93}{0.53} & \valstd{33.74}{0.45} \\ 
\addlinespace 
\textbf{Hot Lots 3}                &    On-Time              & \valstdb{51.77\%}{3.71}   & \valstd{44.40\%}{4.44} & \valstd{41.06\%}{3.93} & \valstd{43.32\%}{3.76} & \valstd{44.5\%}{3.43} \\
\small{Theoretical Cycle-Time: $\mathbf{24.75}$}              &    Cycle-Time              & \valstdb{33.56}{0.17}   & \valstd{33.86}{0.19}  & \valstd{34.08}{0.16} & \valstd{33.94}{0.15} & \valstd{33.87}{0.15}       \\ 
\textbf{Hot Lots 4}  &    On-Time              & \valstdb{59.94\%}{3.45}     & \valstd{51.94\%}{3.84} & \valstd{50.33\%}{3.4} & \valstd{51.7\%}{3.84} & \valstd{52.64\%}{3.33}\\
\small{Theoretical Cycle-Time: $\mathbf{14.54}$}  &    Cycle-Time              & \valstdb{19.52}{0.11}   & \valstd{19.79}{0.12}  & \valstd{19.86}{0.11}   & \valstd{19.82}{0.13} & \valstd{19.77}{0.10}        \\ 
\addlinespace 
\textbf{Regular Lots 3} &    On-Time              & \valstdb{51.72\%}{2.47}   & \valstd{0\%}{0}    & \valstd{0\%}{0} & \valstd{0\%}{0}  & \valstd{0\%}{0}   \\
\small{Theoretical Cycle-Time: $\mathbf{24.75}$} &    Cycle-Time              & \valstdb{63.71}{1.90}      & \valstd{106.91}{1.59} & \valstd{109.9}{0.97}   & \valstd{87.58}{0.72} & \valstd{86.24}{0.55}    \\ 
\textbf{Regular Lots 4} &    On-Time              & \valstd{38.1\%}{2.04}     & \valstd{0\%}{0}   & \valstdb{69.14\%}{2.49} & \valstd{0\%}{0} & \valstd{0\%}{0}  \\
\small{Theoretical Cycle-Time: $\mathbf{14.54}$} &    Cycle-Time             & \valstd{40.77}{1.7}   & \valstd{53.25}{2.02}  & \valstdb{29.43}{0.25}     & \valstd{51.6}{1.15}  & \valstd{52.52}{0.93}   \\ 
\midrule 
& \textbf{Cost}    & \valstdb{33\,171.37}{1\,655.53}                  & \valstd{39\,813.37}{899.89}         & \valstd{40\,615.84}{1\,660.51} & \valstd{39\,695.29}{1\,341.17}  & \valstd{39\,519.56}{1\,484.84}       \\
  \bottomrule
  \end{tabular}
  }
  \caption{High-Volume/Low-Mix (\textit{HV/LM})}\label{apx:HVLM6month}

\end{subtable}

\begin{subtable}[h]{\textwidth}
  \centering

\resizebox{\textwidth}{!}{
  \begin{tabular}{l l r@{$\displaystyle \,\pm\,$}l r@{$\displaystyle \,\pm\,$}l r@{$\displaystyle \,\pm\,$}l r@{$\displaystyle \,\pm\,$}l r@{$\displaystyle \,\pm\,$}l}
  \toprule
  \multicolumn{2}{c}{\textbf{Lot Type}}      & \multicolumn{2}{c}{\textbf{Ours}}                & \multicolumn{2}{c}{\textbf{Hierarchical SPT}} & \multicolumn{2}{c}{\textbf{Hierarchical SRPT}}  & \multicolumn{2}{c}{\textbf{Hierarchical EDD}}  & \multicolumn{2}{c}{\textbf{Hierarchical LS}}  \\
  \midrule 
\textbf{Super Hot Lots 3} &   On-Time              & \valstd{97.38\%}{4.63}   & \valstdb{98.79\%}{2.45} & \valstd{97.48\%}{4.81}  & \valstd{96.9\%}{5.23} & \valstd{97.46\%}{4.63}\\
\small{Theoretical Cycle-Time: $\mathbf{24.75}$} &    Cycle-Time  & \valstd{33.64}{0.44}    & \valstdb{33.32}{0.35} & \valstd{33.67}{0.44}  & \valstd{33.61}{0.44} & \valstd{33.72}{0.45} \\ 
\addlinespace 
\textbf{Hot Lots 1} &   On-Time              & \valstd{97.08\%}{3.29}   & \valstd{96.66\%}{2.09} & \valstdb{97.16\%}{2.5} & \valstd{97\%}{3.27} & \valstd{96.76\%}{3.65}\\
\small{Theoretical Cycle-Time: $\mathbf{21.75}$} &    Cycle-Time & \valstd{29.81}{0.29}    & \valstdb{29.65}{0.19} & \valstd{29.82}{0.28} & \valstd{29.81}{0.23} & \valstd{29.93}{0.28}  \\ 
\textbf{Hot Lots 2} &   On-Time              & \valstdb{98.65\%}{2.2}   & \valstd{97.83\%}{2.01} & \valstd{98.81\%}{1.86} & \valstd{98.38\%}{2.26} & \valstd{98.77\%}{2.03}\\
\small{Theoretical Cycle-Time: $\mathbf{23.3}$} &    Cycle-Time  & \valstdb{31.5}{0.24}   & \valstd{31.6}{0.23}  & \valstd{31.68}{0.22} & \valstd{31.65}{0.27} & \valstd{31.67}{0.23}\\ 
\textbf{Hot Lots 3} &   On-Time              & \valstdb{97.93\%}{2.55}   & \valstd{96.83\%}{2.04} & \valstd{98.18\%}{2.53} & \valstd{97.67\%}{2.94} & \valstd{97.99\%}{2.47} \\
\small{Theoretical Cycle-Time: $\mathbf{24.75}$} & Cycle-Time  & \valstdb{33.56}{0.32}    & \valstd{33.8}{0.23} & \valstd{33.82}{0.29} & \valstd{33.81}{0.28} & \valstd{33.81}{0.29} \\ 
\textbf{Hot Lots 4} &   On-Time              & \valstd{96.62\%}{3.55}    & \valstd{95.9\%}{2.43} & \valstd{96.4\%}{3.53} & \valstdb{97.81\%}{2.88} & \valstd{97.18\%}{3.26} \\
\small{Theoretical Cycle-Time: $\mathbf{14.54}$} & Cycle-Time  & \valstd{19.87}{0.22}    & \valstdb{19.79}{0.16} & \valstd{19.9}{0.23} & \valstd{19.8}{0.19} & \valstd{19.83}{0.24} \\ 
\textbf{Hot Lots 5} &   On-Time              & \valstd{94.17\%}{3.03}  & \valstd{94.02\%}{2.42} & \valstd{93.86\%}{3.14}  & \valstd{94.26\%}{3.14} & \valstdb{94.92\%}{3.32} \\
\small{Theoretical Cycle-Time: $\mathbf{10.10}$} & Cycle-Time  & \valstd{14.09}{0.17}  & \valstdb{14}{0.15} & \valstdb{14}{0.17} & \valstd{14}{0.16} & \valstd{13.96}{0.17} \\ 
\textbf{Hot Lots 6} &   On-Time              & \valstd{96.42\%}{2.62}    & \valstd{96.63\%}{1.91} & \valstd{96.92\%}{2.49} & \valstdb{97.23\%}{2.88} & \valstdb{97.23\%}{2.38}\\
\small{Theoretical Cycle-Time: $\mathbf{12.95}$} & Cycle-Time  & \valstd{17.47}{0.16}   & \valstd{17.46}{0.13} & \valstd{17.45}{0.17} & \valstd{17.47}{0.19} & \valstdb{17.42}{0.19}\\ 
\textbf{Hot Lots 7} &   On-Time              & \valstd{95.22\%}{3.68}  & \valstdb{95.71\%}{2.61} & \valstd{95.85\%}{3.52} & \valstd{94.49\%}{3.39} & \valstd{94.68\%}{3.57}\\
\small{Theoretical Cycle-Time: $\mathbf{15.46}$} & Cycle-Time  & \valstd{21.02}{0.21}   & \valstdb{20.9}{0.16} & \valstd{20.91}{0.25}  & \valstd{20.99}{0.19} & \valstd{21}{0.21} \\ 
\textbf{Hot Lots 8} &   On-Time              & \valstd{92.41\%}{3.71}  & \valstdb{93.7\%}{2.6} & \valstd{93.23\%}{2.86} & \valstd{93.01\%}{3.64} & \valstd{93.31\%}{3.08}\\
\small{Theoretical Cycle-Time: $\mathbf{16.02}$} & Cycle-Time  & \valstd{22.21}{0.21}  & \valstdb{20.9}{0.16} & \valstd{22.09}{0.22} & \valstd{22.14}{0.23} & \valstd{22.05}{0.2}\\ 
\textbf{Hot Lots 9} &   On-Time              & \valstd{95.96\%}{3.28}  & \valstd{95.96\%}{2.26}  & \valstdb{96.96\%}{3} & \valstd{96.65\%}{3.01} & \valstd{96.27\%}{3.05}\\
\small{Theoretical Cycle-Time: $\mathbf{16.95}$} & Cycle-Time  & \valstd{23.16}{0.2} & \valstdb{22.92}{0.14} & \valstd{23}{0.23} & \valstd{23.07}{0.21} & \valstd{23.04}{0.19} \\ 
\textbf{Hot Lots 10} &   On-Time              & \valstd{95.88\%}{4.1}   & \valstd{96.1\%}{1.92} & \valstd{97.16\%}{3.41} & \valstd{96.8\%}{3.25} & \valstdb{97.76\%}{2.67} \\
\small{Theoretical Cycle-Time: $\mathbf{17.32}$} & Cycle-Time  & \valstd{24.01}{0.28}  & \valstd{23.92}{0.2} & \valstd{23.92}{0.21} & \valstdb{23.91}{0.23} & \valstdb{23.91}{0.23}\\ 
\addlinespace 
 \textbf{Regular Lots 1} &   On-Time              & \valstdb{65.80\%}{1.77}   & \valstd{0\%}{0}& \valstd{0\%}{0} & \valstd{0\%}{0} & \valstd{0\%}{0}\\
 \small{Theoretical Cycle-Time: $\mathbf{21.75}$} &    Cycle-Time       & \valstdb{50.77}{1.67}  & \valstd{92.04}{1.88} & \valstd{70.87}{1.46} & \valstd{62.59}{0.39} & \valstd{62.79}{0.42}\\ 
 \textbf{Regular Lots 2} &   On-Time              & \valstdb{69.58\%}{1.80}   & \valstd{0\%}{0} & \valstd{0\%}{0} & \valstd{0\%}{0} & \valstd{0\%}{0}\\
 \small{Theoretical Cycle-Time: $\mathbf{23.3}$} &    Cycle-Time    & \valstdb{49.31}{1.26}  & \valstd{96.8}{1.78} & \valstd{107.63}{2.4} & \valstd{67.27}{0.42}& \valstd{65.96}{0.43}\\
 \textbf{Regular Lots 3} &   On-Time              & \valstdb{58.29\%}{2.42} & \valstd{0\%}{0} & \valstd{0\%}{0}& \valstd{0\%}{0} & \valstd{0\%}{0}\\
 \small{Theoretical Cycle-Time: $\mathbf{24.75}$} & Cycle-Time  & \valstdb{61.85}{2.38}  & \valstd{101.23}{1.86} & \valstd{122.26}{4.63} & \valstd{71.34}{0.42} & \valstd{69.87}{0.44}\\ 
 \textbf{Regular Lots 4} &   On-Time              & \valstd{46.02\%}{3.7}    & \valstd{4.32\%}{3.87}& \valstdb{74.31\%}{4.11}& \valstd{0\%}{0} & \valstd{0\%}{0}\\
 \small{Theoretical Cycle-Time: $\mathbf{14.54}$} & Cycle-Time  & \valstd{40.85}{2.58}  & \valstd{47}{3.37}  & \valstdb{30.43}{0.53}  & \valstd{45.57}{0.48} & \valstd{45.4}{0.45} \\
 \textbf{Regular Lots 5} &   On-Time              & \valstd{24.3\%}{2.12}    & \valstd{0.18\%}{0.41} & \valstdb{90.2\%}{0.49}& \valstd{0.28\%}{0.69} & \valstd{0\%}{0} \\
 \small{Theoretical Cycle-Time: $\mathbf{10.10}$} & Cycle-Time  & \valstd{41.57}{2.48}   & \valstd{58.7}{2.14} & \valstdb{16.4}{0.09} & \valstd{33.75}{0.54} & \valstd{35.7}{0.34} \\
 \textbf{Regular Lots 6} &   On-Time              & \valstd{58.18\%}{2.70}    & \valstd{0\%}{0} & \valstdb{69.13\%}{5.29}& \valstd{0\%}{0} & \valstd{0\%}{0} \\
 \small{Theoretical Cycle-Time: $\mathbf{12.95}$} & Cycle-Time  & \valstd{33.55}{1.57}  & \valstd{71.1}{1.91}  & \valstdb{29.08}{0.57} & \valstd{43.48}{0.42} & \valstd{43.27}{0.43} \\
 \textbf{Regular Lots 7} &   On-Time              & \valstd{65.29\%}{2.24}   & \valstd{0\%}{0} & \valstdb{81.55\%}{1.28}& \valstd{0\%}{0} & \valstd{0\%}{0} \\
 \small{Theoretical Cycle-Time: $\mathbf{15.46}$} & Cycle-Time  & \valstd{36.7}{1.42}  & \valstd{74.96}{1.87}  & \valstdb{30.02}{0.34}  & \valstd{47.02}{0.46}  & \valstd{47.65}{0.46}  \\
 \textbf{Regular Lots 8} &   On-Time              & \valstdb{67.04\%}{1.83}  & \valstd{0\%}{0} & \valstd{59.89\%}{3.36}& \valstd{0\%}{0} & \valstd{0\%}{0}\\
 \small{Theoretical Cycle-Time: $\mathbf{16.02}$} & Cycle-Time  & \valstd{38.43}{1.50}  & \valstd{77.84}{1.86}  & \valstdb{37.59}{0.56} & \valstd{50.74}{0.29}  & \valstd{50.78}{0.31}   \\
 \textbf{Regular Lots 9} &   On-Time              & \valstd{53.25\%}{2.66}   & \valstd{0\%}{0}  & \valstdb{68.75\%}{2.16}& \valstd{0\%}{0} & \valstd{0\%}{0}\\
 \small{Theoretical Cycle-Time: $\mathbf{16.95}$} & Cycle-Time  & \valstd{44.48}{1.60}   & \valstd{79.7}{1.89}  & \valstdb{36.07}{0.48} & \valstd{51.24}{0.41}  & \valstd{52}{0.43}     \\
 \textbf{Regular Lots 10} &   On-Time              & \valstdb{56.97\%}{2.39}  & \valstd{0\%}{0}  & \valstd{53.87\%}{5.15} & \valstd{0\%}{0} & \valstd{0\%}{0}\\
 \small{Theoretical Cycle-Time: $\mathbf{17.32}$} & Cycle-Time  & \valstd{43.95}{1.67}   & \valstd{82.14}{1.79}  & \valstdb{39.93}{0.68} & \valstd{53.09}{0.42} & \valstd{53.23}{0.43} \\
\midrule 
& \textbf{Cost}   & \valstdb{94\,349.38}{2\,716.51}   & \valstd{141\,538.20}{2\,390.37}  & \valstd{115\,084.5}{2\,961.95}  & \valstd{111\,275.17}{2\,067.71}  & \valstd{109\,030.26}{2\,290.72}    \\
  \bottomrule

  \end{tabular}
  }
  \caption{Low-Volume/High-Mix (\textit{LV/HM})}\label{apx:LVHM6month}

\end{subtable}  
\caption{Detailed results of additional dispatching rules on the \textit{SMT2020} testbed for the \textit{HV/LM} and \textit{LV/HM} scenarios with $6$ months of simulation.
In addition to the original dispatching heuristics previously reported, we compare against the Shortest Processing Time (SPT), Shortest Remaining Process Time (SRPT), Earliest Due Date (EDD), and Least Slack (LS) dispatching rules.
We observe similar results as with the original dispatching rules.
}\label{apx:full_6months}
\end{table*}

\begin{table*}[ht]

  \begin{subtable}[h]{\linewidth}
    \centering

  \resizebox{0.85\textwidth}{!}{
    \begin{tabular}{l l r@{$\displaystyle \,\pm\,$}l r@{$\displaystyle \,\pm\,$}l r@{$\displaystyle \,\pm\,$}l}
\toprule
\multicolumn{2}{c}{\textbf{Lot Type}}       & \multicolumn{2}{c}{\textbf{Ours}}                & \multicolumn{2}{c}{\textbf{Hierarchical CR}}              & \multicolumn{2}{c}{\textbf{Hierarchical FIFO}} \\
\midrule \textbf{Super Hot Lots 3} &   On-Time              & \valstd{44.58\%}{9.54}   & \valstdb{46.21\%}{9.46}  & \valstd{40.21\%}{8.03}  \\
\small{Theoretical Cycle-Time: $\mathbf{24.75}$} &    Cycle-Time              & \valstdb{33.73}{0.38}   & \valstd{33.77}{0.37}  & \valstd{33.98}{0.11}  \\ 
\addlinespace 
\textbf{Hot Lots 3}                &    On-Time              & \valstdb{48.68\%}{2.77}   & \valstd{43.99\%}{2.86}  & \valstd{41.22\%}{3.09}    \\
\small{Theoretical Cycle-Time: $\mathbf{24.75}$}              &    Cycle-Time              & \valstdb{33.75}{0.12}   & \valstd{33.91}{0.12}  & \valstd{34.12}{0.13}            \\ 
\textbf{Hot Lots 4}  &    On-Time              & \valstdb{57.04\%}{2.68}   & \valstd{50.77\%}{2.97}  & \valstd{50.47\%}{2.57}   \\
\small{Theoretical Cycle-Time: $\mathbf{14.54}$}  &    Cycle-Time              & \valstdb{19.65}{0.07}   & \valstd{19.84}{0.08}  & \valstd{19.86}{0.08}             \\ 
\addlinespace 
\textbf{Regular Lots 3} &    On-Time              & \valstdb{49.35\%}{1.61}   & \valstd{0\%}{0}  & \valstd{0\%}{0}          \\
\small{Theoretical Cycle-Time: $\mathbf{24.75}$} &    Cycle-Time              & \valstdb{79.91}{4.89}   & \valstd{100.52}{1.06}  & \valstd{98.27}{1.13}            \\ 
\textbf{Regular Lots 4} &    On-Time              & \valstdb{38.72\%}{1.69}   & \valstd{0\%}{0}  & \valstd{0\%}{0}        \\
\small{Theoretical Cycle-Time: $\mathbf{14.54}$} &    Cycle-Time             & \valstdb{47.8}{2.26}   & \valstd{53.16}{1.05}  & \valstd{55.67}{0.68}            \\ 
\midrule 
& \textbf{Cost}    & \valstdb{39\,775.41}{1\,036.54}                  & \valstd{46\,402.56}{636.07}              & \valstd{41\,079.57}{743.56}        \\
\bottomrule
\end{tabular}
}

\caption{High-Volume/Low-Mix (\textit{HV/LM})}\label{tbl:HVLM_1year}

\end{subtable}

\begin{subtable}[h]{\linewidth}
  \centering

\resizebox{0.85\textwidth}{!}{
  \begin{tabular}{l l r@{$\displaystyle \,\pm\,$}l r@{$\displaystyle \,\pm\,$}l r@{$\displaystyle \,\pm\,$}l}
  \toprule
  \multicolumn{2}{c}{\textbf{Lot Type}}       & \multicolumn{2}{c}{\textbf{Ours}}                & \multicolumn{2}{c}{\textbf{Hierarchical CR}}              & \multicolumn{2}{c}{\textbf{Hierarchical FIFO}} \\
  \midrule 
\textbf{Super Hot Lots 3} &   On-Time              & \valstdb{98.16\%}{2.92}   & \valstd{97.9\%}{3.24}  & \valstd{96.69\%}{3.46}  \\
\small{Theoretical Cycle-Time: $\mathbf{24.75}$} &    Cycle-Time              & \valstdb{33.57}{0.33}   & \valstd{33.73}{0.42}  & \valstd{33.75}{0.37}  \\ 
\addlinespace 
\textbf{Hot Lots 1} &   On-Time              & \valstd{96.01\%}{2.52}   & \valstdb{96.43\%}{2.26}  & \valstd{95.06\%}{2.43}  \\
\small{Theoretical Cycle-Time: $\mathbf{21.75}$} &    Cycle-Time              & \valstdb{29.92}{0.21}   & \valstd{29.9}{0.27}  & \valstd{30.05}{0.2}  \\ 
\textbf{Hot Lots 2} &   On-Time              & \valstdb{98.88\%}{1.87}   & \valstd{98.49\%}{1.56}  & \valstd{96.9\%}{2.44}  \\
\small{Theoretical Cycle-Time: $\mathbf{23.3}$} &    Cycle-Time              & \valstdb{31.66}{0.18}   & \valstd{31.66}{0.26}  & \valstd{31.93}{0.21}  \\ 
\textbf{Hot Lots 3} &   On-Time              & \valstdb{96.88\%}{2.50}   & \valstd{97.60\%}{1.83}  & \valstd{95.63\%}{2.52}  \\
\small{Theoretical Cycle-Time: $\mathbf{24.75}$} & Cycle-Time  & \valstdb{33.83}{0.24}   & \valstd{33.8}{0.18}  & \valstd{34.14}{0.21}  \\ 
\textbf{Hot Lots 4} &   On-Time              & \valstd{94.86\%}{2.65}   & \valstdb{96.38\%}{2.47}  & \valstd{94.65\%}{2.65}  \\
\small{Theoretical Cycle-Time: $\mathbf{14.54}$} & Cycle-Time  & \valstdb{19.93}{0.16}   & \valstd{19.92}{0.16}  & \valstd{20.03}{0.14}  \\ 
\textbf{Hot Lots 5} &   On-Time              & \valstdb{93.79\%}{2.19}   & \valstd{94.21\%}{2.1}  & \valstd{93.98\%}{2.43}  \\
\small{Theoretical Cycle-Time: $\mathbf{10.10}$} & Cycle-Time  & \valstdb{14.14}{0.14}   & \valstd{14.13}{0.13}  & \valstd{14.10}{0.12}  \\ 
\textbf{Hot Lots 6} &   On-Time              & \valstd{95.95\%}{2.07}   & \valstdb{97.03\%}{1.86}  & \valstd{96.04\%}{2.28}  \\
\small{Theoretical Cycle-Time: $\mathbf{12.95}$} & Cycle-Time  & \valstdb{17.56}{0.13}   & \valstd{17.58}{0.15}  & \valstd{17.58}{0.14}  \\ 
\textbf{Hot Lots 7} &   On-Time              & \valstdb{94\%}{2.28}   & \valstd{93.73\%}{2.33}  & \valstd{93.4\%}{2.51}  \\
\small{Theoretical Cycle-Time: $\mathbf{15.46}$} & Cycle-Time  & \valstdb{21.15}{0.16}   & \valstd{21.12}{0.16}  & \valstd{21.21}{0.14}  \\ 
\textbf{Hot Lots 8} &   On-Time              & \valstdb{92.5\%}{2.87}   & \valstd{92.89\%}{2.71}  & \valstd{91.47\%}{3.1}  \\
\small{Theoretical Cycle-Time: $\mathbf{16.02}$} & Cycle-Time  & \valstdb{22.24}{0.15}   & \valstd{22.23}{0.15}  & \valstd{22.36}{0.18}  \\ 
\textbf{Hot Lots 9} &   On-Time              & \valstd{95.25\%}{2.53}   & \valstdb{96.02\%}{2.46}  & \valstd{94.83\%}{2.4}  \\
\small{Theoretical Cycle-Time: $\mathbf{16.95}$} & Cycle-Time  & \valstd{23.1}{0.15}   & \valstdb{23}{0.16}  & \valstd{23.2}{0.16}  \\ 
\textbf{Hot Lots 10} &   On-Time              & \valstd{95.53\%}{2.78}   & \valstdb{96.89\%}{2.38}  & \valstd{95.19\%}{2.3}  \\
\small{Theoretical Cycle-Time: $\mathbf{17.32}$} & Cycle-Time  & \valstd{24.04}{0.19}   & \valstdb{23.96}{0.19}  & \valstd{24.12}{0.18}  \\ 
\addlinespace 
 \textbf{Regular Lots 1} &   On-Time              & \valstdb{64.61\%}{1.62}   & \valstd{0\%}{0}  & \valstd{0\%}{0}  \\
 \small{Theoretical Cycle-Time: $\mathbf{21.75}$} &    Cycle-Time       & \valstdb{59.61}{3.06}   & \valstd{65.97}{0.48}  & \valstd{71.88}{0.87}  \\ 
 \textbf{Regular Lots 2} &   On-Time              & \valstdb{68.92\%}{1.8}   & \valstd{0\%}{0}  & \valstd{0\%}{0}  \\
 \small{Theoretical Cycle-Time: $\mathbf{23.3}$} &    Cycle-Time    & \valstdb{59.17}{2.74}   & \valstd{76.14}{0.66}  & \valstd{73.9}{0.87}  \\ 
 \textbf{Regular Lots 3} &   On-Time              & \valstdb{56.42\%}{1.88}   & \valstd{0\%}{0}  & \valstd{0\%}{0}  \\
 \small{Theoretical Cycle-Time: $\mathbf{24.75}$} & Cycle-Time  & \valstdb{73.06}{3.08}   & \valstd{80.45}{0.67}  & \valstd{81.77}{0.92}  \\ 
 \textbf{Regular Lots 4} &   On-Time              & \valstdb{40.24\%}{2.34}   & \valstd{0\%}{0}  & \valstd{0\%}{0}  \\
 \small{Theoretical Cycle-Time: $\mathbf{14.54}$} & Cycle-Time  & \valstdb{62.05}{3.03}   & \valstd{48.62}{0.45}  & \valstd{46.7}{0.59}  \\ 
 \textbf{Regular Lots 5} &   On-Time              & \valstdb{29.11\%}{2}   & \valstd{0\%}{0}  & \valstd{0\%}{0}  \\
 \small{Theoretical Cycle-Time: $\mathbf{10.10}$} & Cycle-Time  & \valstd{50.06}{2}   & \valstdb{33.68}{0.27}  & \valstd{34.2}{0.5}  \\ 
 \textbf{Regular Lots 6} &   On-Time              & \valstdb{67.51\%}{2.08}   & \valstd{0\%}{0}  & \valstd{0.29\%}{0.45}  \\
 \small{Theoretical Cycle-Time: $\mathbf{12.95}$} & Cycle-Time  & \valstdb{32.13}{1.57}   & \valstd{47.9}{0.5}  & \valstd{40.64}{0.54}  \\ 
 \textbf{Regular Lots 7} &   On-Time              & \valstdb{65.05\%}{1.71}   & \valstd{0\%}{0}  & \valstd{0\%}{0}  \\
 \small{Theoretical Cycle-Time: $\mathbf{15.46}$} & Cycle-Time  & \valstdb{42.96}{1.77}   & \valstd{48.66}{0.44}  & \valstd{49.08}{0.61}  \\ 
 \textbf{Regular Lots 8} &   On-Time              & \valstdb{70.46\%}{1.49}   & \valstd{0\%}{0}  & \valstd{0\%}{0}  \\
 \small{Theoretical Cycle-Time: $\mathbf{16.02}$} & Cycle-Time  & \valstdb{41.92}{1.66}   & \valstd{56.55}{0.51}  & \valstd{52.9}{0.64}  \\ 
 \textbf{Regular Lots 9} &   On-Time              & \valstdb{54.84\%}{2.66}   & \valstd{0\%}{0}  & \valstd{0\%}{0}  \\
 \small{Theoretical Cycle-Time: $\mathbf{16.95}$} & Cycle-Time  & \valstdb{49.16}{1.56}   & \valstd{51.5}{0.37}  & \valstd{53.47}{0.64}  \\ 
 \textbf{Regular Lots 10} &   On-Time              & \valstdb{54.66\%}{1.95}   & \valstd{0\%}{0}  & \valstd{0\%}{0}  \\
 \small{Theoretical Cycle-Time: $\mathbf{17.32}$} & Cycle-Time  & \valstdb{53.85}{1.93}   & \valstd{54.96}{0.44}  & \valstd{55.78}{0.69}  \\ 
\midrule 
& \textbf{Cost}    & \valstdb{109\,643.74}{1\,848.16}                  & \valstd{123\,895.88}{1\,840.75}              & \valstd{111\,977.53}{1\,780.09}        \\
  \bottomrule

  \end{tabular}

  }

  \caption{Low-Volume/High-Mix (\textit{LV/HM})}\label{tbl:LVHM_1year}

\end{subtable}
\caption{Detailed results on the \textit{SMT2020} testbed for the \textit{HV/LM} and \textit{LV/HM} scenarios with a $1$ year simulation horizon.
The dispatching agents, which were trained over a shorter horizon of $6$ months, are the same as in Tables \ref{tbl:LVHM/HVLM}--\ref{apx:full_6months}.
}\label{tbl:HVLM_LVHM_1year}
\end{table*}

\end{document}